\pdfoutput=1

\documentclass[11pt]{article}

\usepackage[]{acl}

\usepackage{times}
\usepackage{latexsym}

\usepackage[T1]{fontenc}

\usepackage[utf8]{inputenc}

\usepackage{microtype}
\usepackage{multirow}

\usepackage{makecell}
\usepackage{adjustbox}
\usepackage{booktabs}

\usepackage{graphicx}

\usepackage{amsfonts}
\usepackage{mathabx}

\usepackage{xspace}

\title{\SUPERBsg: Enhanced Speech processing Universal PERformance Benchmark for Semantic and Generative Capabilities}

\author{
    Hsiang-Sheng Tsai$^{1}$\thanks{$^*$Equal contribution.} , Heng-Jui Chang$^{1*}$, Wen-Chin Huang$^{2*}$, Zili Huang$^{3*}$, \\
    \bf{Kushal Lakhotia$^{4*}$, Shu-wen Yang$^1$, Shuyan Dong$^4$, Andy T. Liu$^1$, Cheng-I  Jeff Lai$^5$,}\\
    \bf{Jiatong Shi$^6$, Xuankai Chang$^6$, Phil Hall$^7$, Hsuan-Jui Chen$^1$,}\\
    \bf{Shang-Wen Li$^4$, Shinji Watanabe$^6$, Abdelrahman Mohamed$^4$, Hung-yi Lee$^1$} \\ 
    \\
    $^1$National Taiwan University, Taiwan \\
    $^2$Nagoya University, Japan \\
    $^3$Johns Hopkins University, USA \\
    $^4$Meta AI, USA \\
    $^5$Massachusetts Institute of Technology, USA \\
    $^6$Carnegie Mellon University, USA \\
    $^7$LXT\\
    \\
    \texttt{\{r09922024, b06901020\}@ntu.edu.tw} \\
    \texttt{wen.chinhuang@g.sp.m.is.nagoya-u.ac.jp} \\
    \texttt{hzili1@jhu.edu, kushall@fb.com} \\
    \texttt{shangwel@fb.com, shinjiw@ieee.org} \\
    \texttt{abdo@fb.com, hungyilee@ntu.edu.tw}
}

\newcommand{\SUPERB}{\texttt{SUPERB}\xspace}
\newcommand{\SUPERBsg}{\texttt{SUPERB-SG}\xspace}

\begin{document}

\maketitle
\begin{abstract}
Transfer learning has proven to be crucial in advancing the state of speech and natural language processing research in recent years. In speech, a model pre-trained by self-supervised learning transfers remarkably well on multiple tasks. However, the lack of a consistent evaluation methodology is limiting towards a holistic understanding of the efficacy of such models. \SUPERB was a step towards introducing a common benchmark to evaluate pre-trained models across various speech tasks. In this paper, we introduce \SUPERBsg, a new benchmark focused on evaluating the semantic and generative capabilities of pre-trained models by increasing task diversity and difficulty over \SUPERB. We use a lightweight methodology to test the robustness of representations learned by pre-trained models under shifts in data domain and quality across different types of tasks. It entails freezing pre-trained model parameters, only using simple task-specific trainable heads. The goal is to be inclusive of all researchers, and encourage efficient use of computational resources. We also show that the task diversity of \SUPERBsg coupled with limited task supervision is an effective recipe for evaluating the generalizability of model representation.
\end{abstract}

\section{Introduction}
\label{sec:introduction}
Transfer learning is a paradigm in machine learning that has been very effective for natural language processing (NLP) \cite{elmo,bert,liu2019roberta,albert,dong2019unified,yang2019xlnet,raffel2020exploring,lewis2019bart,conneau2020unsupervised}, and speech processing~\cite{cpc,modified_cpc,apc1,wav2vec,wav2vec2,hsu2021hubert,mockingjay,tera,pase+,decoar,decoar2}. Self-supervised learning (SSL) is the main driver of this paradigm, an effective and scalable way to learn high-level representation of language that transfers to a variety of tasks. SSL entails learning from the input or some perturbation of it without the need for labelled data. This has unlocked the usage of large amounts of cheaply available unlabelled data. It lends naturally to neural network models that have been shown to possess impressive scaling characteristics such that it is often enough to increase the model and data sizes to improve downstream performance \cite{hestness2017deep,shazeer2017outrageously,jozefowicz2016exploring,mahajan2018exploring,radford2019language}.


Speech signal consists of acoustic, linguistic, prosodic, and speaker characteristics. SSL algorithms in speech must be evaluated in their ability to produce representations that are useful for tasks that demand understanding of linguistic, speaker, and prosodic elements of spoken language as well as high-level semantics. Researchers have used auto-regressive, contrastive, discriminative and multi-task learning objectives to pre-train models, and have investigated their capabilities across tasks like phoneme recognition~\cite{cpc,apc1}, automatic speech recognition (ASR)~\cite{tera,wav2vec,decoar2,pase+,hsu2021hubert,distilhubert}, speaker verification~\cite{sv-wav2vec2}, speaker identification~\cite{apc1,mockingjay}, emotion recognition~\cite{macary2021use}, speech translation~\cite{apc1}, voice conversion~\cite{lin2020fragmentvc, vqw2v-vc}, spoken language understanding~\cite{lai2020semi}, and text-to-speech~\cite{ssl-for-tts}. 
However, the methodologies in such studies vary in the use of datasets, fine-tuning strategies and task-specific model architectures.
To bridge this gap, \SUPERB~\cite{yang2021superb} introduced a standardized benchmark of 10 speech tasks to compare 13 pre-trained models and a Log Mel-Filterbank baseline. It studied the models' performance in tasks focusing on linguistic (phoneme recognition and automatic speech recognition, keyword spotting and query by example), shallow semantic (intent classification and slot filling), speaker (speaker identification, speaker verification and speaker diarization), and prosodic (emotion recognition) characteristics.


In this paper, we introduce \SUPERBsg, a benchmark with 5 new tasks, which are speech translation, out-of-domain ASR, voice conversion, speech separation, and speech enhancement, with an emphasis on evaluating the semantic and generative capabilities of pre-trained models that require high-level representations to capture linguistic, semantic, and speaker characteristics. These tasks go beyond speech recognition by focusing on various other aspects that are essential to building intelligent speech interfaces. Further, we show that while SSL models achieve close to state-of-the-art performance on many tasks, there isn't one model that outperforms all others, and that a simple Log Mel-Filterbank can perform competitively on some tasks. We also demonstrate the robustness of our methodology with an ablation study over different task-specific model architectures and data sizes.

The introduction of these new tasks of varying difficulty takes us closer to a more comprehensive unified standard speech benchmark. We hope that this will motivate the development of more powerful, generalizable, and reusable pre-trained models to democratize the advancement of speech research. To facilitate this, 
we released the codes\footnote{\label{toolkit}https://github.com/s3prl/s3prl:
Tasks in \SUPERBsg are open-sourced and reproducible in the S3PRL toolkit which supports benchmarking the most existing and customized pre-trained models.
}
and integrated the tasks with the \SUPERB benchmark.

\begin{figure}[t]
    \centering
    \includegraphics[width=1\columnwidth]{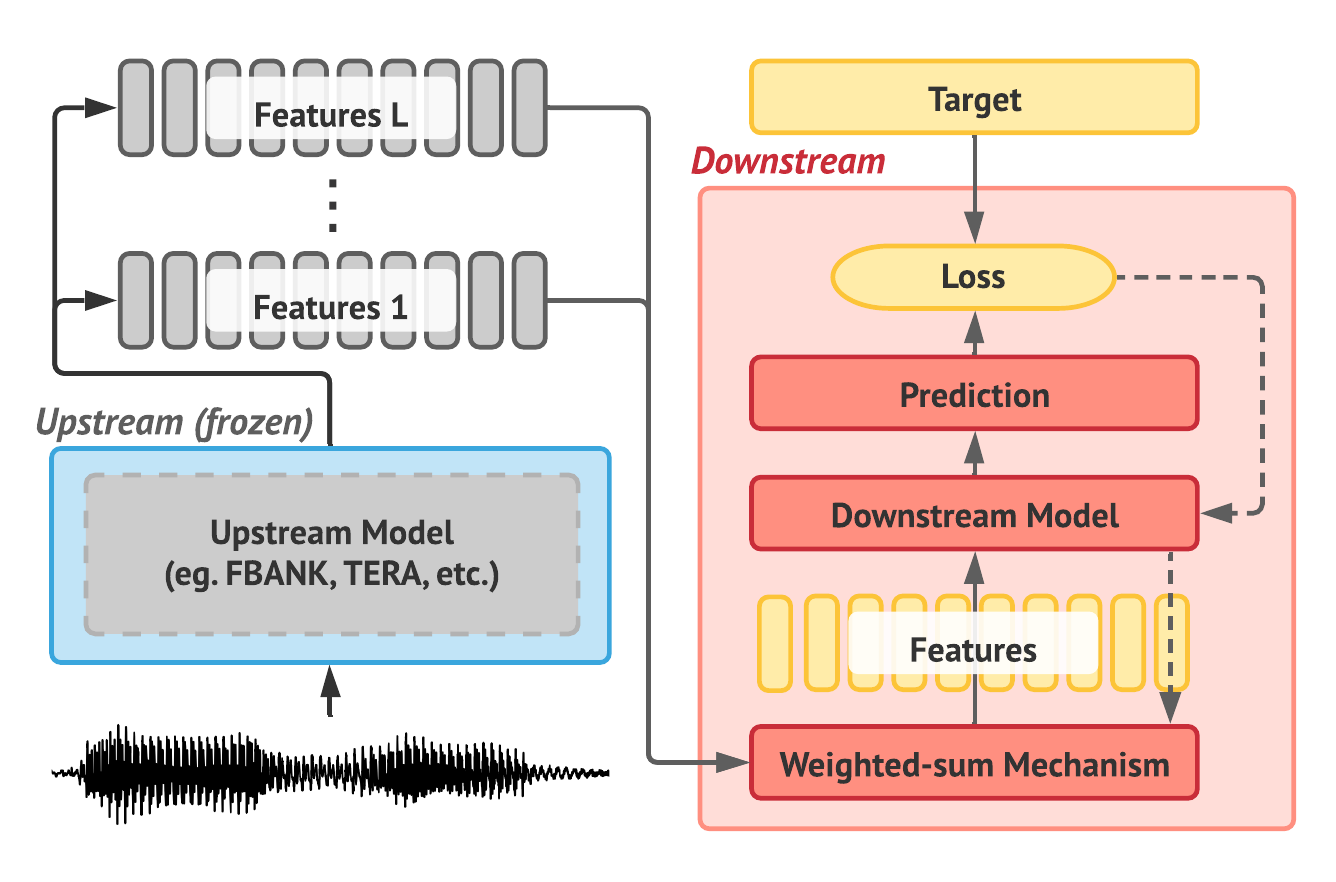}
    \caption{Illustration of the detailed training procedure. A trainable weighted-sum mechanism is used to  summarize all layers’ representations into a sequence of vectors and then taken by downstream model as input. Upstream is frozen through the whole process. Dashed arrow ($\dashrightarrow$) is used to indicate the flow of gradient when back propagating.}
    \label{fig:diagram}
\end{figure}

\section{Related Work}

As more powerful SSL models are proposed with promising performance on various tasks, researchers continually try to find extensive evaluation methods to assess model performance, and wish to holistically understand the capability of the learned representations in these models. 

\SUPERB~\cite{yang2021superb} is a framework to benchmark the SSL models on 10 speech tasks by learning task-specific prediction heads on top of the frozen shared SSL models.
Although the tasks in \SUPERB span across different domains, most of them are simple classification problems, or only require utilization of shallow semantics.
In contrast, we focus on harder semantic and generative tasks.

Another recently proposed benchmark is the LeBenchmark~\cite{evain2021lebenchmark}, investigating the performance of SSL models trained on French data with three semantic tasks.
However, they only consider wav2vec 2.0~\cite{wav2vec2} with different architectures as their upstream models (i.e., networks pre-trained with SSL).
Here, we evaluate a diverse set of SSL models, and offer a more comprehensive analysis.

The Zero Resource Speech Benchmark 2021~\cite{zerospeech} introduces unsupervised speech processing tasks, particularly the spoken language modeling problem.
They evaluate the SSL models via zero-shot probings at four linguistic levels.
While their benchmark task is specific for certain domain, we use various tasks to evaluate different aspects of SSL models.

The HEAR 2021 Challenge\footnote{https://neuralaudio.ai/hear2021-holistic-evaluation-of-audio-representations.html} aims to develop general-purpose audio representation by focusing on audio tasks beyond speech that include sound event detection, speech commands and pitch \& chroma classification. We specifically focus on various aspects of speech processing, thus providing a wide variety of spoken language tasks.

\begin{table*}[t]
\centering
\setlength{\tabcolsep}{3pt}
\begin{adjustbox}{max width=\linewidth}
\begin{tabular}{@{}lcrccccc}
\toprule
Upstream & Network & \#Params & Stride & Input & Corpus & Pretraining & Official Github \\
\midrule

FBANK & - & 0 & 10ms & waveform & - & - & - \\ 
\midrule

PASE+ & SincNet, 7-Conv, 1-QRNN & 7.83M & 10ms & waveform & LS 50 hr & multi-task & santi-pdp / pase \\
\midrule

APC & 3-GRU & 4.11M & 10ms & FBANK & LS 360 hr & F-G & iamyuanchung / APC \\

VQ-APC & 3-GRU & 4.63M & 10ms & FBANK & LS 360 hr & F-G + VQ & iamyuanchung / VQ-APC \\

NPC & 4-Conv, 4-Masked Conv & 19.38M & 10ms & FBANK & LS 360 hr & M-G + VQ & Alexander-H-Liu / NPC \\

Mockingjay & 12-Trans & 85.12M & 10ms & FBANK & LS 360 hr & time M-G & s3prl / s3prl \\

TERA & 3-Trans & 21.33M & 10ms & FBANK & LS 960 hr & time/freq M-G & s3prl / s3prl \\ 

DeCoAR 2.0 & 12-Trans & 89.84M & 10ms & FBANK & LS 960 hr & time M-G + VQ & awslabs / speech-representations \\ 
\midrule

Modified CPC & 5-Conv, 1-LSTM & 1.84M & 10ms & waveform & LL 60k hr & F-C & facebookresearch / CPC\_audio \\

wav2vec & 19-Conv & 32.54M & 10ms & waveform & LS 960 hr & F-C & pytorch / fairseq \\

vq-wav2vec & 20-Conv & 34.15M & 10ms & waveform & LS 960 hr & F-C + VQ & pytorch / fairseq \\

wav2vec 2.0 Base & 7-Conv 12-Trans & 95.04M & 20ms & waveform & LS 960 hr & M-C + VQ & pytorch / fairseq \\

wav2vec 2.0 Large & 7-Conv 24-Trans & 317.38M & 20ms & waveform & LL 60k hr & M-C + VQ & pytorch / fairseq \\

HuBERT Base & 7-Conv 12-Trans & 94.68M & 20ms & waveform & LS 960 hr & M-P + VQ & pytorch / fairseq \\

HuBERT Large & 7-Conv 24-Trans & 316.61M & 20ms & waveform & LL 60k hr & M-P + VQ & pytorch / fairseq \\
\bottomrule
\end{tabular}
\end{adjustbox}
\caption{
Details of the investigated SSL representations. LibriSpeech and LibriLight are denoted as LS and LL, respectively. For the pretraining methods, we abbreviate "vector quantization" as VQ, "future" as F, "masked" as M, "generation" as G, "contrastive discrimination" as C, and "token prediction/classification" as P. Parameters for both pretraining and inference are counted.
}
\label{tab:upstreams}
\end{table*}

\section{SUPERB-SG}
\label{sec:superbsg}

\subsection{Tasks and Datasets}
\label{sec:tasks}

This section introduces the tasks in \SUPERBsg, including why we choose these tasks and how we design the task-specific heads for fine-tuning. Following \SUPERB's methodology, we use a lightweight fine-tuning approach wherein we freeze the pre-trained model parameters and only keep the task-specific head's parameters trainable. 
This setting serves the dual purpose of evaluating the robustness as well as the generalizability of the speech representations, and provides a resource-efficient way of fine-tuning the models that is inclusive of participants with constrained compute resources.
We call the pre-trained model as upstream model and the task-specific heads as downstream model. We now discuss the newly added tasks in \SUPERBsg in the following sub-sections.

\subsubsection{Speech Translation}
Speech translation (ST) involves translating the acoustic speech signals in the source language into the words in the target language. We use it to evaluate the semantic capability of SSL models, and how they benefit the translation task. We use the CoVoST2 En$\rightarrow$De~\cite{covost2} dataset
(CC0 Licensed)
with their official train, validation, and test splits while removing all the samples containing "REMOVE", resulting in 425.8, 25.9 and 24.5 hours respectively. For text, we keep original case, normalize punctuation, and build character vocabulary with 100\% train-set coverage.
We report case-sensitive de-tokenized BLEU using sacreBLEU~\cite{sacrebleu}.
Our downstream model has an encoder-decoder architecture with 3 layers of Transformers~\cite{46201} each
with hidden dimension of 512.
A convolutional sub-sampler is used to reduce the sequence length of the input before feeding it to the encoder. We train our model with label-smoothing using a probability of 0.1. A beam size of 20 is used for inference.

\subsubsection{Out-of-domain ASR}
Although an ASR is included in \SUPERB, it only examines SSL models on read English corpus LibriSpeech~\cite{panayotov2015librispeech}.
Therefore, we introduce out-of-domain ASR (OOD-ASR), which aims to evaluate the models' capabilities across languages, and out-of-domain scenarios.
The OOD-ASR tasks are categorized into cross-lingual and spontaneous speech tasks.
For the cross-lingual tasks, we choose the Mexican Spanish (es), Mandarin (zh), and Arabic (ar) subsets from Common Voice 7.0~\cite{common-voice}
(CC0 Licensed)
containing 21.5, 31.2, and 30.7 hours of training data respectively. The validation set sizes are 1.2 hours, 14.4 hours and 12.24 hours, and the test set sizes are 0.6 hour, 15.3 hours and 12.5 hours for es, zh and ar respectively.
For the spontaneous speech task (spon), we use the Santa Barbara Corpus of Spoken American English (SBCSAE)~\cite{du2000sbcsae} 
(CC BY-ND 3.0 Licensed),
consisting of 60 conversations over different topics spanning 16.7 hours of data. The validation and test set sizes are 1.6 hours and 2.2 hours respectively.
For evaluation, we use word error rate (WER) as the metric except for Mandarin which character error rate (CER) is used. The error rates are averaged across all sub-tasks to offer an overall score.
The ASR model is a 2-layer BLSTM~\cite{hochreiter1997long} with hidden states of 1024 dimension. The training objective is to minimize the Connectionist Temporal Classification (CTC) loss~\cite{ctc}.
During inference, we use CTC greedy decoding without language model re-scoring to simplify the process and to highlight the impact of  the learned acoustic representations.

\subsubsection{Voice Conversion}
For voice conversion (VC), we consider the intra-lingual VC task in VCC2020~\cite{vcc2020} 
(ODbL Licensed)
under the any-to-one (A2O) setting. A2O VC aims to convert speech from any arbitrary speaker into that of a predefined target speaker.  We use the task to evaluate the speaker transferability as well as the generalizability of the SSL models.
We use 60 utterances from the target speaker that spans 5 minutes for training, and 25 utterances for testing that span 2 minutes. No validation set was used.
We use the commonly used mel-cepstrum distortion (MCD), word error rate (WER) and automatic speaker verification (ASV) accept rate from off-the-shelf ASR and ASV models as evaluation metrics.
The downstream model is trained to reconstruct the acoustic feature from the upstream representations in a target-speaker-dependent manner. In the conversion phase, given the representations extracted by the upstream, the model generates the converted acoustic features, which are then sent to a neural vocoder to synthesize the converted waveform. We adopted Tacotron2~\cite{Taco2} as the downstream model, which is an autoregressive network consisting of convolutional and LSTM layers. For the neural vocoder, we used the Hifi-GAN~\cite{hifigan}.
We follow an implementation described in \cite{huang2021s3prl}.

\subsubsection{Speech Separation}
Speech separation (SS) is the task of separating target speech from background interference~\cite{wang2018supervised}. It is an important step in speech processing, especially for noisy and multi-speaker scenarios.
We investigate the speech separation problem on a dataset simulated from LibriSpeech~\cite{cosentino2020librimix} 
(CC BY 4.0 Licensed)
and WHAM!~\cite{wichern2019wham} 
(CC BY-NC 4.0 Licensed)
noise. We use 16kHz version of the dataset containing 2 speakers, and focus on the \textit{mix\_clean} condition. The train and evaluation sets contain 43.3 and 4.2 hours of speech simulated from LibriSpeech's \textit{train-clean-100} and \textit{test-clean}.
This task is used to evaluate the generative capability of SSL models when input is a mixture of acoustic signals. 
We use the scale-invariant signal-to-distortion ratio improvement (SI-SDRi) as the evaluation metric. For the downstream model, we use a 3-layer BLSTM model with dimension of 896 for each direction to predict the short-time Fourier transform (STFT) masks for each speaker, and the predictions are transformed back to the time domain using inverse short-time Fourier transform (iSTFT).
Permutation invariant training (PIT)~\cite{yu2017permutation} is performed to optimize the mean square error between the predicted mask and Ideal Non-negative Phase Sensitive Mask (INPSM)~\cite{erdogan2015phase,kolbaek2017multitalker}. We choose frequency domain method instead of a time domain based method because of the stride size constraint and computational cost.

\begin{table*}[ht!]
\centering
\small
\begin{adjustbox}{max width=\textwidth}
\begin{tabular}{@{}l r c r c rrr c r c rr}
\toprule
\multirow{2}{*}{Upstream} & \multicolumn{1}{c}{ST} && \multicolumn{1}{c}{OOD-ASR} && \multicolumn{3}{c}{VC} && \multicolumn{1}{c}{SS} && \multicolumn{2}{c}{SE} \\
\cmidrule{2-2} \cmidrule{4-4} \cmidrule{6-8} \cmidrule{10-10} \cmidrule{12-13}
& BLEU$\uparrow$ && WER$\downarrow$ && MCD$\downarrow$ & WER$\downarrow$ & ASV$\uparrow$ && SI-SDRi$\uparrow$ && PESQ$\uparrow$ & STOI$\uparrow$ \\ \midrule

FBANK        & 2.32 && 63.58 && 8.47 & 38.3 & 77.25 && 9.23 && 2.55 & 93.6 \\
\midrule
PASE+        & 3.16 && 61.56 && 8.66 & 30.6 & 63.20 && 9.87 && 2.56 & 93.9 \\
\midrule
APC          & 5.95 && 63.12 && 8.05 & 27.2 & 87.25 && 8.92 && 2.56 & 93.4 \\
VQ-APC       & 4.23 && 63.56 && 7.84 & 22.4 & 94.25 && 8.44 && 2.56 & 93.4 \\
NPC          & 4.32 && 61.66 && 7.86 & 30.4 & 94.75 && 8.04 && 2.52 & 93.1 \\
Mockingjay   & 4.45 && 65.27 && 8.29 & 35.1 & 79.75 && 9.29 && 2.53 & 93.4 \\
TERA         & 5.66 && 58.49 && 8.21 & 25.1 & 83.75 && 10.19 && 2.54 & 93.6 \\
DeCoAR 2.0   & 9.94 && 53.62 && 7.83 & 17.1 & 90.75 && 8.54 && 2.47 & 93.2 \\
\midrule
Modified CPC & 4.82 && 62.54 && 8.41 & 26.2 & 71.00 && 10.40 && 2.57 & 93.7 \\
wav2vec      & 6.61 && 55.86 && 7.45 & 10.1 & 98.25 && 9.30 && 2.53 & 93.8 \\
vq-wav2vec   & 5.66 && 60.66 && \textbf{7.08} & 13.4 & \textbf{100.00} && 8.16 && 2.48 & 93.6 \\
wav2vec 2.0 Base  & 14.81 && 46.95 && 7.50 & 10.5 & 98.00 && 9.77 && 2.55 & 93.9 \\
wav2vec 2.0 Large & 12.48 && 44.69 && 7.63 & 15.8 & 97.25 && 10.02 && 2.52 & 94.0 \\
HuBERT Base       & 15.53 && 46.69 && 7.47 & \textbf{8.0} & 98.50 && 9.36 && 2.58 & 93.9 \\
HuBERT Large      & \textbf{20.01} && \textbf{44.08} && 7.22 & 9.0 & 99.25 && \textbf{10.45} && \textbf{2.64} & \textbf{94.2} \\
\bottomrule
\end{tabular}
\end{adjustbox}
\caption{Evaluating various SSL representations on new semantic and generative downstream tasks. $\uparrow$ indicates the higher the better and $\downarrow$ indicates the lower the better. The complete results of OOD-ASR are in Appendix \ref{sec:full_oodasr}.}
\label{tab: Main}
\end{table*}

\subsubsection{Speech Enhancement}
Speech enhancement (SE) is the task of removing background noise from a degraded speech signal, and it aims to improve the perceived quality and intelligibility of the signal. We include this task to evaluate the generative capability under noisy conditions.
In \SUPERBsg, we discuss the speech enhancement problem on the Voicebank-DEMAND~\cite{veaux2013voice} 
(CC BY 4.0 Licensed)
corpus. The train, validation, and test sets contain 8.8, 0.6 and 0.6 hours of speech respectively. 
Our evaluation metrics are Perceptual Evaluation of Speech Quality (PESQ) and Short-Time Objective Intelligibility (STOI). For the downstream model, we follow the mask-based speech enhancement pipeline in~\cite{kolbaek2017multitalker}. A 3-layer BLSTM model similar to the speech separation task is trained to predict the spectral mask for the clean signal.
The mean square error between the predicted mask and INPSM is used as the objective.

\subsection{Self-supervised Models}

We evaluate the tasks on 15 upstream models, which are PASE+ \cite{pase+}, APC \cite{apc1}, VQ-APC \cite{vq_apc}, NPC \cite{npc}, Mockingjay \cite{mockingjay}, TERA \cite{tera}, DeCoAR 2.0 \cite{decoar2}, Modifile CPC \cite{modified_cpc}, wav2vec family~\cite{wav2vec}~\cite{vq_wav2vec}~\cite{wav2vec2} and HuBERT~\cite{hsu2021hubert}. They span across different architectures, sizes and learning objectives. Some models also use vector quantization which has an added benefit of signal compression.
For grounding, we use Log Mel Filterbank as our baseline. The detailed properties of upstream models are shown in Table \ref{tab:upstreams}.



\section{Experimental Setup}
\label{sec:setup}

Following \SUPERB, we fix upstream models parameters for all downstream tasks' training.
We extract the frame-level representations for each hidden layer of the upstream models from raw waveform, and use a trainable task-specific weighted-sum mechanism to summarize all layers' representations into a sequence of vectors.
The summarized representations are then used as the downstream model's input. 
An overview of the training procedure is demonstrated in Figure~\ref{fig:diagram}.
Each experiment is done by one single run with the same seed.
This procedure is consistent for all experiments, offering a fair and simple evaluation strategy for all upstream models.

%

\section{Results and Discussion}


\subsection{Main result}


The results of the upstream models evaluated on \SUPERBsg are shown in Table \ref{tab: Main}.
We only report the averaged WER for OOD-ASR. Full results can be found in Appendix \ref{sec:full_oodasr}.
For speech-to-text tasks (ST and OOD-ASR), wav2vec 2.0 and HuBERT offer competitive results,  while DeCoAR 2.0 shows some improvements.
In speech generation tasks (VC, SS, and SE), FBANK yields comparable or superior performance than some SSL models, especially for those metrics that take the quality of the output signal into account.
For VC, the 3 reported metrics have the same trend for respective models.
Here, vq-wav2vec achieves the best performance on MCD and ASV, while HuBERT performs the best on WER.
For SS, Hubert-Large achieves the best performance, followed by Modified CPC. PASE+, which is pre-trained with denoising tasks, performs better than half the SSL models, but this observation doesn't transfer to the other tasks.
For SE, all upstream models perform comparably. The largest gap is only 0.17 in PESQ and 1.1 in STOI.

Overall, no model outperforms all others on all tasks. However, HuBERT-Large performs most competitively on all downstream tasks, especially those requiring linguistic and semantic signals.

\begin{figure}[t]
    \centering
    \includegraphics[width=\columnwidth]{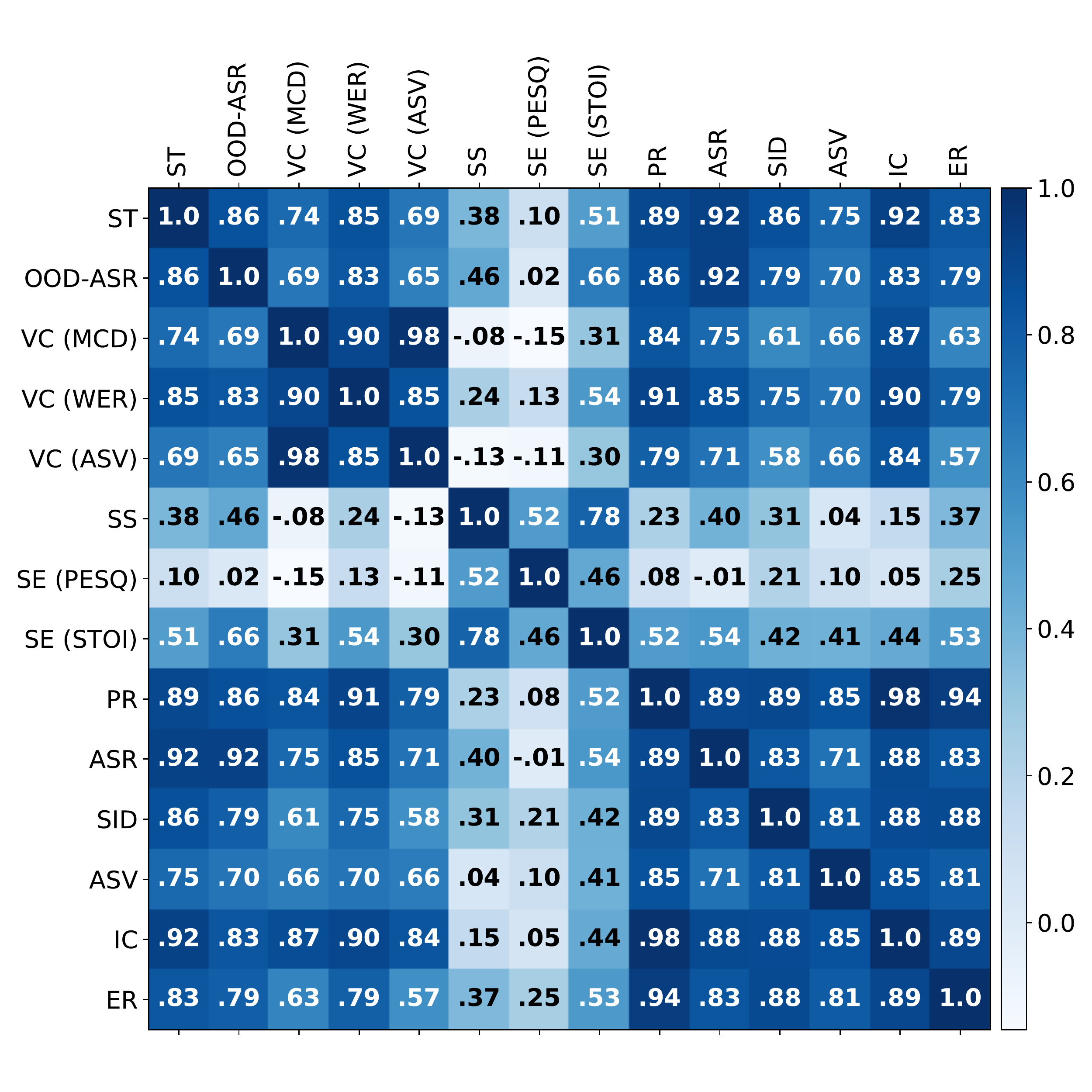}
    \caption{Spearman's $\rho$ between tasks.}
    \label{fig:correlation}
\end{figure}

\begin{figure}[t]
    \centering
    \includegraphics[width=\columnwidth]{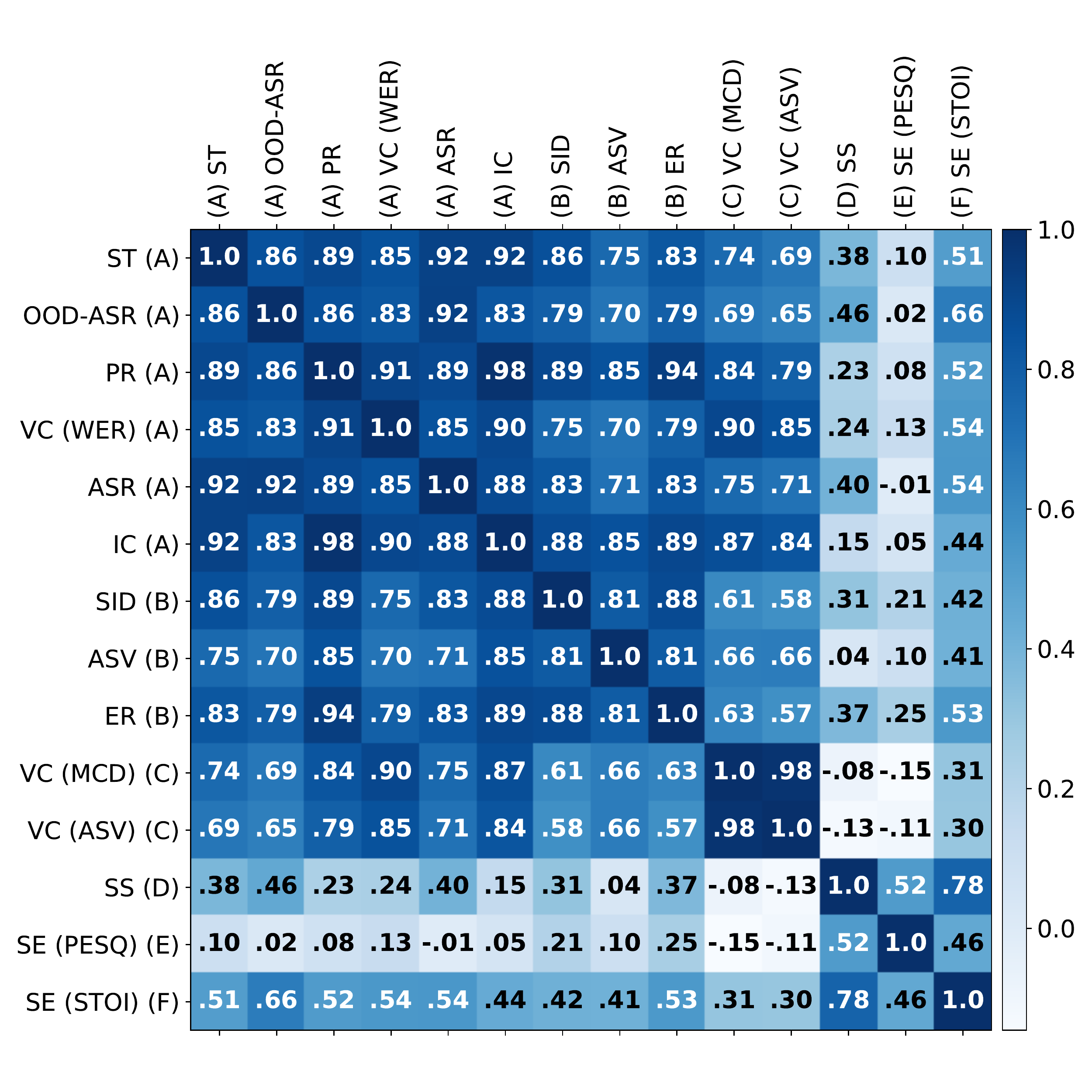}
    \caption{Spearman's $\rho$ between tasks rearranged by clustering result.}
    \label{fig:correlation clustering}
\end{figure}

\subsection{Correlation between tasks}


We analyze the correlations between tasks in \SUPERBsg to understand the similarity between tasks, and verify if the experimental results agree with the common understanding of related tasks based on shared representation they require.


To compute the correlation, we first change all metrics into a higher-better manner.
Then, we compute the Spearman's rank correlation coefficients (Spearman's $\rho$) between all pairs of tasks.
For multiple metrics contained in a single task, such as MCD/WER/ASV in VC as well as PESQ/STOI in SE, we compute each of them separately. 

To make our analysis more representative and generalized to all speech domains, we bring back the six tasks from \SUPERB~\cite{yang2021superb} that are considered representative of the following four domains: 
(\romannumeral 1) Content recognition tasks containing Phoneme Recognition (PR), Automatic Speech Recognition (ASR)
(\romannumeral 2) Speaker identity tasks including Identification (SID), Automatic Speaker Verification (ASV)
(\romannumeral 3) Semantics task which is Intent Classification (IC)
and (\romannumeral 4) Prosodic task which is Emotion Recognition (ER).
Together with the 5 tasks introduced in this paper, we show the results of total 11 downstream tasks with the 14 corresponding metrics in Figure \ref{fig:correlation}.

Overall, results show that all tasks except SS and SE have strong positive correlation among them. 
One possible explanation for SS and SE not showing strong correlation is that the low-level information closely related to audio signals is more critical as they need to reconstruct clean speech from interfering speakers and background noise by estimating the STFT masks.
As a result, high-level information extracted from SSL models has little benefit for these tasks but is helpful for other tasks. As noted earlier, there is only a small gap in performance between FBANK and SSL models.
If we leave SS and SE out, all correlation coefficients are greater than 0.58, showing that the SSL model representations are useful for multiple domains.

Although the Spearman's $\rho$ are large in general in Figure \ref{fig:correlation}, differences between tasks are observable.
Here, we focus on the relation between correlation and similarity of tasks.
We list the most and the least two correlated tasks comparing with ST, OOD-ASR, VC, SS, and SE.
SS and SE are skipped as candidates for for the least correlated tasks since they dominate the results.
For VC, we average the correlation coefficients across the three metrics.
The results are shown in Table~\ref{tab:correlated tasks}.
ST and OOD-ASR are highly correlated with ASR since they both transform speech signals into discrete text tokens.
IC is also correlated with ST since semantic information is required to perform both tasks.
Moreover, ASV and VC are the least correlated tasks since they primarily focus on the speaker information with lesser regard to the semantic content. However, the absolute correlation values are still larger than 0.7.
For VC, the speaker information needs to be removed while the content has to be kept, similar to PR and ASR but different from SID.
SS and SE are correlated with each other and have a much lower correlation with speaker identity and semantics tasks, supporting our assumption. 
Overall, we find that empirically highly-correlated tasks require similar knowledge or understanding ability.

\begin{table}[t]
    \centering
    \small
    \begin{adjustbox}{max width=\columnwidth}
    \begin{tabular}{ccccc}
        \toprule
        Tasks & Top 2 && Last 2 \\
        \midrule
        ST & \makecell{ASR \\ (0.92)} \makecell{IC \\ (0.92) } && \makecell{ASV  \\ (0.75)} \makecell{VC \\ (0.76)}  \\
        \midrule
        OOD-ASR & \makecell{ASR \\ (0.92)} \makecell{PR \\ (0.86) }  && \makecell{ASV  \\ (0.70)} \makecell{VC \\ (0.72)}  \\
        \midrule
        VC & \makecell{PR \\ (0.84)} \makecell{ASR \\ (0.77) } && \makecell{SID  \\ (0.64)} \makecell{ER \\ (0.66)}  \\ 
        \midrule 
        SS & \makecell{SE \\ (0.65)} \makecell{OOD-ASR \\ (0.46) }  && \makecell{VC  \\ (0.01)} \makecell{ASV \\ (0.04)}  \\
        \midrule
        SE & \makecell{SS \\ (0.65)} \makecell{ER \\ (0.39) }  && \makecell{VC  \\ (0.17)} \makecell{IC \\ (0.25)}  \\
        \bottomrule
    \end{tabular}
    \end{adjustbox}
    \caption{Top 2 and last 2 tasks correlated with the five \SUPERBsg tasks ranked by Spearman's $\rho$.}
    \label{tab:correlated tasks}
\end{table}

\begin{table}[t]
    \centering
    \small
    \begin{adjustbox}{max width=\columnwidth}
    \begin{tabular}{cc}
        \toprule
        Cluster & Metrics \\ 
        \midrule
        A & \makecell{ST, OOD-ASR, PR\\ VC (WER), ASR, IC} \\
        \midrule
        B & SID, ASV, ER \\
        \midrule
        C & VC (MCD), VC (ASV) \\
        \midrule
        D & SS  \\
        \midrule
        E & SE (PESQ) \\
        \midrule
        F & SE (STOI) \\
        \toprule
    \end{tabular}
    \end{adjustbox}
    \caption{K-means clustering result based on the correlation between each downstream tasks.}
    \label{tab: correlation cluster}
\end{table}

To give a broader view of our correlation results, we further cluster the downstream tasks by their correlation with each other using K-means. In this way, all the tasks are considered simultaneously, and the grouping is driven automatically by the empirical correlation results.
If more than one metric are used in a downstream task, we cluster them independently.
The clustering results are shown in Table \ref{tab: correlation cluster} and a rearranged correlation map is shown in Figure \ref{fig:correlation clustering}.
The result shows that the clusters of the tasks align with our empirical knowledge.
Cluster A includes tasks that require content information, while tasks in cluster B are more sensitive to speaker and prosodic features.
Cluster C contains metrics MCD and ASV of VC, which are used to evaluate the signal quality and the rates of speaker transfer.
It is worth noting that WER in VC belongs to cluster A, showing that it is more similar to content-related tasks.
Furthermore, clusters D, E, and F each contain one of the metrics in SS and SE, aligning with our assumption that these tasks utilize different types of information compared to other tasks.


With the analysis of the correlation between tasks, we empirically confirm the reliability of the results, and show that we increase the heterogeneity among speech tasks over \SUPERB. We further discover shared properties between tasks with clustering, and the result is aligned with our common understanding of related tasks.

\begin{table*}[t]
    \small
    \centering

    \begin{tabular}{@{}ccclcclcc}
        \toprule
        \multirow{2}{*}{Architecture} & \multicolumn{2}{c}{ST} && \multicolumn{2}{c}{OOD-ASR} && \multicolumn{2}{c}{SS} \\
        \cmidrule{2-3} \cmidrule{5-6} \cmidrule{8-9}
        & architecture & \#params && architecture & \#params && architecture & \#params \\
        \midrule
        \textit{default} & \makecell{3-layer encoder \\ 3-layer decoder \\Transformer \\ (dim 512)} & 28.8M && \makecell{2-layer BLSTM \\ (dim 1024)} & 53.4M && \makecell{3-layer BLSTM \\ (dim 896)} & 51.4M \\
        \midrule
        \textit{small} & \makecell{no encoder \\ 1-layer decoder \\Transformer \\ (dim 512)} & \makecell{10.9M \\ ($\times$ 0.38)} && \makecell{1-layer BLSTM \\ (dim 1024)} & \makecell{24.1M \\ ($\times$ 0.45)} && \makecell{2-layer BLSTM \\ (dim 768)} & \makecell{24.4M \\ ($\times$ 0.47)} \\
        \midrule
        \textit{large} & \makecell{12-layer encoder \\ 6-layer decoder \\Transformer \\ (dim 512)} & \makecell{ 69.8M \\ ($\times$ 2.42)} && \makecell{4-layer BLSTM \\ (dim 1024)} & \makecell{ 112.2M \\ ($\times$ 2.10)} && \makecell{4-layer BLSTM \\ (dim 1152)} & \makecell{ 114.50M \\ ($\times$ 2.23)} \\
        \bottomrule
    \end{tabular}
    \caption{A detailed comparison of downstream model architectures. We report the number of trainable parameters when using TERA as upstream model while minor difference (< 10\%) exists due to different upstream dimensions. For OOD-ASR, we average values across all sub-tasks since sub-tasks have different vocabulary sizes.}
    \label{tab:ablation model arch}
    
\end{table*}

\begin{table}[t]
    \centering
    \small
    \begin{adjustbox}{max width=0.95\columnwidth}
    \begin{tabular}{@{}l r @{}l@{} r @{}l@{} r}
        \toprule
        \multirow{2}{*}{Upstream} & \multicolumn{1}{c}{ST} & ~~ & \multicolumn{1}{c}{OOD-ASR} & ~~ & \multicolumn{1}{c}{SS} \\
        \cmidrule{2-2} \cmidrule{4-4} \cmidrule{6-6}
        & \multicolumn{1}{c}{BLEU$\uparrow$} && \multicolumn{1}{c}{WER$\downarrow$} && \multicolumn{1}{c}{SI-SDRi$\uparrow$} \\
        \midrule
        \textit{default} \\
        \midrule
        \quad FBANK            & 2.32 && 63.58 && 9.23 \\
        \quad TERA             & 5.66 && 58.49 && 10.19 \\
        \quad Modified CPC     & 4.82 && 62.54 && \textbf{10.40}\\
        \quad wav2vec 2.0 Base & 14.81 && 46.95 && 9.77 \\
        \quad HuBERT Base      & \textbf{15.53} && \textbf{46.69} && 9.36 \\
        \midrule
        \textit{small} \\
        \midrule
        \quad FBANK            & 0.58 && 70.86 && 8.19 \\
        \quad TERA             & 1.84 && 64.80 && 9.20 \\
        \quad Modified CPC     & 1.44 && 67.83 && \textbf{9.56} \\
        \quad wav2vec 2.0 Base & 8.55 && 50.75 && 8.83 \\
        \quad HuBERT Base      & \textbf{9.24} && \textbf{50.32} && 8.73 \\
        \midrule
        \textit{large} \\
        \midrule
        \quad FBANK            & 3.02 && 60.49 && 9.77 \\
        \quad TERA             & 6.64 && 57.95 && ($\blacktriangleup$) \textbf{10.87} \\
        \quad Modified CPC     & 4.56 && 59.73 && ($\blacktriangledown$) 10.61 \\
        \quad wav2vec 2.0 Base & 16.81 && ($\blacktriangleup$) \textbf{45.61} && 9.86 \\
        \quad HuBERT Base      & \textbf{17.59} && ($\blacktriangledown$) 45.78 && 9.83 \\
        \bottomrule
    \end{tabular}
    
    \end{adjustbox}
    \caption{Results on SS, ST, OOD-ASR when using different architectures. $\blacktriangleup$ and $\blacktriangledown$ are used to denote the rank changing. The complete results of OOD-ASR are in Appendix \ref{sec:full_oodasr}.}
    \label{tab:abaltion model result}
\end{table}

\subsection{Robustness of SUPERB-SG}
\label{sec:robustness}

To study the impact of downstream model architecture and the data sizes used in \SUPERBsg we evaluate the robustness of \SUPERBsg with variations in downstream model as well as training data size, and show that our conclusions still hold true.

We choose ST, OOD-ASR and SS as the downstream tasks for evaluation with an aim to cover semantic, content recognition, and generative task types. For the upstream models, FBANK, TERA, CPC, wav2vec 2.0 Base and HuBERT Base are used to cover different SSL algorithms.

\subsubsection{Downstream model}
\label{sec:downstream}
For each task, 2 additional downstream architectures are created by modifying the number of layers and the hidden dimensions compared to our default setting. We create \textit{small} and \textit{large} models that are roughly the half and twice of \textit{default} in terms of the number of trainable parameters.
A detailed comparison of the downstream architectures is shown in Table \ref{tab:ablation model arch}.
The results are shown in Table \ref{tab:abaltion model result}.

We show that the ranking of the upstream models is almost fixed when the model sizes are varied.
As expected, the \textit{small} architecture has worse performance than \textit{default}, while \textit{large} has better.
Moreover, the scores causing the change in ranking are negligible, e.g., TERA/CPC in SS and wav2vec 2.0 Base/HuBERT Base in OOD-ASR with \textit{large}.
The results show that the relative performance achieved by different upstream models is agnostic to the downstream architecture, confirming the robustness of the framework used in \SUPERBsg.

\subsubsection{Training data size}
\begin{table}[t]
    \centering
    \begin{adjustbox}{max width=0.95\columnwidth}
    \begin{tabular}{@{}lrrrrrr}
        \toprule
        \multirow{2}{*}{Partition} & \multicolumn{1}{c}{\multirow{2}{*}{ST}} & \multicolumn{4}{c}{OOD-ASR} & \multicolumn{1}{c}{\multirow{2}{*}{SS}} \\
        \cmidrule{3-6}
        & & \multicolumn{1}{c}{es} & \multicolumn{1}{c}{zh} & \multicolumn{1}{c}{ar} & \multicolumn{1}{c}{spon} & \\
        \midrule
        Train \\
        \quad \textit{100\%} & 425.80 & 21.44 & 31.05 & 30.39 & 11.43 & 43.27 \\
        \quad \textit{10\%} & 42.58 & 2.15 & 3.11 & 3.04 & 1.14 & 4.34 \\
        \quad \textit{5\%} & 25.91 & 1.07 & 1.56 & 1.52 & 0.57 & 2.17 \\
        \quad \textit{1\%} & 4.26 & 0.22 & 0.31 & 0.31 & 0.12 & 0.43 \\
        \midrule
        Dev & 25.91 & 1.19 & 14.41 & 12.24 & 1.59 & 1.52 \\ 
        \midrule
        Test & 24.51 & 0.62 & 15.32 & 12.46 & 2.15 & 4.19 \\
        \bottomrule
    \end{tabular}
    \end{adjustbox}
    \caption{Hours of data in pseudo datasets. }
    \label{tab:ablation data statistics}
\end{table}
\begin{table}[t]
    \centering
    \begin{adjustbox}{max width=0.95\columnwidth}
    \begin{tabular}{@{}l r @{}l@{} r @{}l@{} r}
        \toprule
        \multirow{2}{*}{Upstream} & \multicolumn{1}{c}{ST} & ~~ & \multicolumn{1}{c}{OOD-ASR} & ~~ & \multicolumn{1}{c}{SS} \\
        \cmidrule{2-2} \cmidrule{4-4} \cmidrule{6-6}
        & \multicolumn{1}{c}{BLEU$\uparrow$} && \multicolumn{1}{c}{WER$\downarrow$} && \multicolumn{1}{c}{SI-SDRi$\uparrow$} \\
        \midrule
        \textit{100\%} \\
        \midrule
        \quad FBANK            & 2.32 && 63.58 && 9.23 \\
        \quad TERA             & 5.66 && 58.49 && 10.19 \\
        \quad Modified CPC     & 4.82 && 62.54 && \textbf{10.40}\\
        \quad wav2vec 2.0 Base & 14.81 && 46.95 && 9.77 \\
        \quad HuBERT Base      & \textbf{15.53} && \textbf{46.69} && 9.36 \\
        \midrule
        \textit{10\%} \\
        \midrule
        \quad FBANK            & 0.46 && 85.39 && 5.65 \\
        \quad TERA             & ($\blacktriangledown$) 0.88 && 80.32 && ($\blacktriangleup$) \textbf{6.72} \\
        \quad Modified CPC     & ($\blacktriangleup$) 1.30 && 85.32 && ($\blacktriangledown$) 6.59 \\
        \quad wav2vec 2.0 Base & 5.04 && 63.85 && 6.45 \\
        \quad HuBERT Base      & \textbf{5.57} && \textbf{63.43} && 6.13 \\
        \midrule
        \textit{5\%} \\
        \midrule
        \quad FBANK            & 0.27 && 89.70 && 4.52 \\
        \quad TERA             & 0.44 && 86.95 && ($\blacktriangleup$ 1) \textbf{5.59} \\
        \quad Modified CPC     & 0.37 && 87.97 && ($\blacktriangledown$ 3) 4.95 \\
        \quad wav2vec 2.0 Base & 2.91 && 69.88 && ($\blacktriangleup$ 1) 5.36 \\
        \quad HuBERT Base      & \textbf{3.35} && \textbf{69.33} && ($\blacktriangleup$ 1) 5.03 \\
        \midrule
        \textit{1\%} \\
        \midrule
        \quad FBANK            & 0.03 && 99.53 && 2.29 \\
        \quad TERA             & 0.04 && 98.31 && 3.24 \\
        \quad Modified CPC     & 0.03 && 98.37 && ($\blacktriangledown$ 3) 2.87\\
        \quad wav2vec 2.0 Base & 0.33 && 92.46 && ($\blacktriangleup$ 2) \textbf{3.34} \\
        \quad HuBERT Base      & \textbf{0.38} && \textbf{92.17} && ($\blacktriangleup$ 1) 3.01 \\
        \bottomrule
    \end{tabular}
    \end{adjustbox}
    \caption{Results on ST, OOD-ASR and SS when using different amount of training data. $\blacktriangleup$ and $\blacktriangledown$ are used to denote the rank changing. The complete results of OOD-ASR are in Appendix \ref{sec:full_oodasr}.}
    \label{tab:abaltion data result}
\end{table}

To study the effect of data size, we create 3 pseudo datasets per task by sub-sampling 10\%, 5\% and 1\% from the original training set while fixing the validation and test sets. The statistics of the datasets are shown in Table \ref{tab:ablation data statistics}, and the results are in Table \ref{tab:abaltion data result}.

The ranking of the upstream models remains almost the same for 10\% of training data.
When that is further reduced to 5\%, there is a change in ranking in SS due to a performance drop in Modified CPC.
Excluding Modified CPC, the ranking is still fixed showing that the relative performance of the upstream models is agnostic to data size. 

Furthermore, when using only 1\% of training data, most of the SSL models fail on the 3 downstream tasks.
This phenomenon is caused by insufficient task-specific knowledge due to limited training data size.
Although SSL models learn high-level representations from the unlabeled speech signal, acquisition of task-specific knowledge such as translingual ability in ST, text-level token mapping in OOD-ASR, and mask prediction in SS, requires non-trivial supervision. 

We note that fewer training examples speeds training up but sacrifices the evaluation quality.
Overall, we show the robustness of \SUPERBsg to variations in data size even when the training data is reduced to 5\%, showing the reliability of the benchmark. 

\section{Conclusion}
We introduce \SUPERBsg, a set of 5 new tasks that include speech translation, out-of-domain ASR, voice conversion, speech separation, and speech enhancement to evaluate the deep semantic and generative capabilities of SSL models. We evaluate 15 SSL models, and do a comprehensive analysis of the task correlations to demonstrate the reliability of our methodology. We test and confirm the robustness of \SUPERBsg in terms of the downstream model architecture as well as the training data size.
The latest introduction of the semantic and generative tasks increases the diversity and difficulty of \SUPERB, which can boost a more comprehensive understanding of the capability of various SSL models' representations, and help researchers discover the hidden properties of SSL techniques in development.

We have open-sourced all the codes\textsuperscript{\ref{toolkit}} 
and released a challenge\footnote{\label{leaderboard}https://superbbenchmark.org}
to encourage further research of SSL in speech. We welcome the community to participate and advance the research frontier together.

\section*{Ethics}
    This work fully adheres to the ACL code of ethics. For more details, we provide a checklist in Appendix \ref{sec: checklist}.
\bibliography{main_acl}
\bibliographystyle{acl_natbib}

\clearpage
\appendix

\section{Complete Out-of-domain ASR Results}
\label{sec:full_oodasr}

Here, we provide complete results of OOD-ASR tasks, as shown in Tables \ref{tab:oodasr_all}, \ref{tab:oodasr_model_size}, \ref{tab:oodasr_data_size}.
All upstream models used in this paper are trained with English speech data, but we are also interested in multilingual pre-trained models in OOD-ASR.
Therefore, we evaluate the wav2vec 2.0 XLSR model on the OOD-ASR tasks, as shown in the last row of Table \ref{tab:oodasr_all}.
XLSR has identical architecture as wav2vec 2.0 Large, but is trained with 56k hours of speech including 53 different languages.
The pre-training data of XLSR cover our cross-lingual tasks' training data.
As expected, using multilingual data improves OOD-ASR tasks and achieves the best performance among all upstream models.

\begin{table}[h]
\centering
\begin{adjustbox}{max width=\columnwidth}
\small
\begin{tabular}{@{}lccccc}
\toprule
\multirow{2}{*}{Upstream} & es & zh & ar & spon &  \\ 
\cmidrule{2-5}
 & WER$\downarrow$ & CER$\downarrow$ & WER$\downarrow$ & WER$\downarrow$ & AVG \\
\midrule
FBANK & 54.03 & 35.44 & 72.07 & 92.78 & 63.58 \\
\midrule
PASE+ & 52.11 & 35.52 & 70.47 & 88.15 & 61.56 \\
\midrule
APC & 55.23 & 36.38 & 70.79 & 90.07 & 63.12 \\
VQ-APC & 55.32 & 37.06 & 71.56 & 90.29 & 63.56 \\
NPC & 51.07 & 35.85 & 69.87 & 89.86 & 61.66 \\
Mockingjay & 58.11 & 38.13 & 73.57 & 91.27 & 65.27 \\
TERA & 48.67 & 32.21 & 66.18 & 86.89 & 58.49 \\
Modified CPC & 54.37 & 36.22 & 68.94 & 90.61 & 62.54 \\
DeCoAR 2.0 & 43.18 & 28.77 & 61.00 & 81.53 & 53.62 \\
wav2vec & 46.16 & 31.69 & 60.85 & 84.72 & 55.86 \\
vq-wav2vec & 52.02 & 36.55 & 66.19 & 87.89 & 60.66 \\
wav2vec 2.0 Base & 37.85 & 26.44 & 55.95 & 67.55 & 46.95 \\
wav2vec 2.0 Large & 35.75$^\dagger$ & 25.07$^\dagger$ & 54.29$^\dagger$ & \textbf{63.64}$^\dagger$ & 44.69 \\
HuBERT Base & 37.15 & 26.23 & 54.94 & 68.41 & 46.69 \\
HuBERT Large & \textbf{30.90} & \textbf{23.73}$^\dagger$ & \textbf{50.60}$^\ddagger$ & 71.09$^\ddagger$ & \textbf{44.08} \\
\midrule
wav2vec 2.0 XLSR & 26.90$^\dagger$ & 22.97$^\dagger$ & 49.63$^\dagger$ & 63.05$^\dagger$ & 40.64$^\dagger$ \\
\bottomrule
\end{tabular}
\end{adjustbox}
\caption{Results of OOD-ASR tasks, where spon denotes spontaneous speech.
$^\dagger$ Normalized across  dimensionality of representation to stabilize training and ensure convergence.
$^\ddagger$ Uses linear warmup of learning rates in the first 8k steps to stabilize training and ensure convergence.}
\label{tab:oodasr_all}
\end{table}

\begin{table}[t]
\centering
\begin{adjustbox}{max width=\columnwidth}
\small
\begin{tabular}{@{}lccccc}
\toprule
\multirow{2}{*}{Upstream} & es & zh & ar & spon &  \\ 
\cmidrule{2-5}
 & WER$\downarrow$ & CER$\downarrow$ & WER$\downarrow$ & WER$\downarrow$ & AVG \\
\midrule
\textit{default}\\
\midrule
\quad FBANK & 54.03 & 35.44 & 72.07 & 92.78 & 63.58 \\
\quad TERA & 48.67 & 32.21 & 66.18 & 86.89 & 58.49 \\
\quad Modified CPC & 54.37 & 36.22 & 68.94 & 90.61 & 62.54 \\
\quad wav2vec 2.0 Base & 37.85 & 26.44 & 55.95 & \textbf{67.55} & 46.95 \\
\quad HuBERT Base & \textbf{37.15} & \textbf{26.23} & \textbf{54.94} & 68.41 & \textbf{46.69} \\
\midrule
\textit{small}\\
\midrule
\quad FBANK & 63.86 & 41.97 & 80.30 & 97.30 & 70.86 \\
\quad TERA & 57.13 & 37.66 & 73.92 & 90.49 & 64.80 \\
\quad Modified CPC & 60.81 & 41.47 & 76.45 & 92.59 & 67.83 \\
\quad wav2vec 2.0 Base & 41.84 & 30.22 & 61.72 & \textbf{69.23} & 50.75 \\
\quad HuBERT Base & \textbf{41.45} & \textbf{29.68} & \textbf{59.93} & 70.21 & \textbf{50.32} \\
\midrule
\textit{large} \\
\midrule
\quad FBANK & 46.39 & 37.71 & 65.35 & 92.52 & 60.49 \\
\quad TERA & 45.41 & 37.40 & 64.48 & 84.53 & 57.95 \\
\quad Modified CPC & 48.70 & 35.16 & 69.15 & 85.93 & 59.73 \\
\quad wav2vec 2.0 Base & 34.02 & 27.60 & 54.10 & \textbf{66.73} & \textbf{45.61} \\
\quad HuBERT Base & \textbf{33.91} & \textbf{27.22} & \textbf{53.43} & 68.57 & 45.78 \\
\bottomrule
\end{tabular}
\end{adjustbox}
\caption{Complete results of OOD-ASR tasks with different model sizes.}
\label{tab:oodasr_model_size}
\end{table}

\begin{table}[t!]
\centering
\begin{adjustbox}{max width=\columnwidth}
\small
\begin{tabular}{@{}lccccc}
\toprule
\multirow{2}{*}{Upstream} & es & zh & ar & spon &  \\ 
\cmidrule{2-5}
 & WER$\downarrow$ & CER$\downarrow$ & WER$\downarrow$ & WER$\downarrow$ & AVG \\
\midrule
\textit{100\%}\\
\midrule
\quad FBANK & 54.03 & 35.44 & 72.07 & 92.78 & 63.58 \\
\quad TERA & 48.67 & 32.21 & 66.18 & 86.89 & 58.49 \\
\quad Modified CPC & 54.37 & 36.22 & 68.94 & 90.61 & 62.54 \\
\quad wav2vec 2.0 Base & 37.85 & 26.44 & 55.95 & \textbf{67.55} & 46.95 \\
\quad HuBERT Base & \textbf{37.15} & \textbf{26.23} & \textbf{54.94} & 68.41 & \textbf{46.69} \\
\midrule
\textit{10\%} \\
\midrule
\quad FBANK            & 84.82 & 62.97 & 93.27 & 100.49 & 85.39 \\
\quad TERA             & 76.44 & 58.54 & 88.49 & 97.79 & 80.32 \\
\quad Modified CPC     & 83.84 & 64.78 & 91.20 & 101.44 & 85.32 \\
\quad wav2vec 2.0 Base & 61.26 & 43.50 & 72.98 & \textbf{77.65} & 63.85 \\
\quad HuBERT Base      & \textbf{58.08} & \textbf{42.94} & \textbf{72.78} & 79.94 & \textbf{63.43} \\
\midrule
\textit{5\%} \\
\midrule
\quad FBANK            & 89.48 & 71.99 & 96.69 & 100.65 & 89.70 \\
\quad TERA             & 83.98 & 71.04 & 93.15 & 99.62 & 86.95 \\
\quad Modified CPC     & 88.61 & 67.61 & 95.71 & 99.93 & 87.97 \\
\quad wav2vec 2.0 Base & 67.09 & \textbf{50.58} & 78.53 & \textbf{83.33} & 69.88 \\
\quad HuBERT Base      & \textbf{66.29} & 50.72 & \textbf{76.59} & 83.74 & \textbf{69.33} \\
\midrule
\textit{1\%} \\
\midrule
\quad FBANK & 96.79 & 96.73 & 99.85 & 104.73 & 99.53 \\
\quad TERA & 94.73 & 98.82 & 99.77 & 99.93 & 98.31 \\
\quad Modified CPC & 95.93 & 97.94 & 99.80 & 99.84 & 98.37 \\
\quad wav2vec 2.0 Base & \textbf{82.00} & 94.38 & 92.41 & \textbf{101.06} & 92.46 \\
\quad HuBERT Base & 82.36 & \textbf{94.34} & \textbf{90.37} & 101.60 & \textbf{92.17} \\
\bottomrule
\end{tabular}
\end{adjustbox}
\caption{Complete results of OOD-ASR tasks with different data sizes.}
\label{tab:oodasr_data_size}
\end{table}

\section{Responsible NLP Research Checklist}

\label{sec: checklist}

Here we answer the ethics questions to show our ethics statement.

\subsection{Did you discuss the \textit{limitations} of your work?}
Yes, we discussed the constrains on the frozen upstreams and simple task specific heads in abstract and Section \ref{sec:superbsg}.

\subsection{Did you discuss any potential \textit{risks} of your work?}
Yes, in Section \ref{sec:robustness}, we discussed about the risks of the unstable benchmark results, and we showed the robustness of \SUPERBsg.

\subsection{Do the abstract and introduction summarize the paper’s main claims?}
Yes, the paper's main claims are summarized in abstract and Section \ref{sec:introduction}.

\subsection{Did you use or create \textit{scientific artifacts}?}
Yes, we used public datasets and pre-trained models mentioned in Section \ref{sec:superbsg}.

\subsubsection{Did you cite the creators of artifacts you used?}
Yes, we cited those artifacts properly in Section \ref{sec:superbsg}.

\subsubsection{Did you discuss the \textit{license or terms} for use and/or distribution of any artifacts?}
Yes, the licenses of the artifacts are clearly indicated in Section \ref{sec:superbsg}.

\subsubsection{Did you discuss if your use of existing artifact(s) was consistent with their \textit{intended use}, provided that it was specified? For the artifacts you create, do you specify intended use and whether that is compatible with the original access conditions (in particular, derivatives of data accessed for research purposes should not be used outside of research contexts)?}
Yes, we use the official implementations of the upstream models in Table \ref{tab:upstreams} and followed their public API to access the models. For the datasets, we also follow their licenses.

\subsubsection{Did you discuss the steps taken to check whether the data that was collected/used contains any \textit{information that names or uniquely identifies individual people} or \textit{offensive content}, and the steps taken to protect / anonymize it?}
No, there were no data collection involved in this work. We used the widely-used public datasets and followed the common data preprocessing steps.

\subsubsection{Did you provide documentation of the artifacts, e.g., coverage of domains, languages, and linguistic phenomena, demographic groups represented, etc.?}
Yes, the properties of the artifacts were indicated in Section \ref{sec:superbsg}.
    
\subsubsection{Did you report relevant statistics like the number of examples, details of train/test/dev splits, etc. for the data that you used/created?}
Yes, the relevant statistics were reported in Section \ref{sec:superbsg}.

\subsection{Did you run \textit{computational experiments}?} 
Yes.

\subsubsection{Did you report the \textit{number of parameters} in the models used, the \textit{total computational budget} (e.g., GPU hours), and \textit{computing infrastructure} used?}

We reported the number of the parameters in Section \ref{sec:downstream}.
The computational budget and computing infrastructures are reported in Table \ref{tab:training time}.
\begin{table}[t]
\centering
\begin{adjustbox}{max width=\columnwidth}
\small
\begin{tabular}{@{}lccccc}
\toprule
 & ST & OOD-ASR & VC & SS & SE \\
 \midrule
 Steps & 32k & 500k & 10k & 150k & 150k \\
 Time & 25hr & 36hr & 4hr & 48hr & 72hr \\
 GPU & 3090 & V100 & 3090 & 1080 Ti & 1080 Ti \\
 \bottomrule
\end{tabular}
\end{adjustbox}
\caption{Training steps, time and GPU devices used by each task when using HuBERT Base as upstream. NVIDIA ReForce RTX 3090, NVIDIA Tesla V100 and NVIDIA GeForce GTX 1080 Ti are denoted as 3090, V100 and 1080 Ti respectively.}
\label{tab:training time}
\end{table}

\subsubsection{Did you discuss the experimental setup, including \textit{hyperparameter search} and \textit{best-found hyperparameter} values?}
No, we didn't do the hyperparameter searching in a unified way. Some hyperparameters came from the official implementation or related works and some were searched by ourselves. However, the hyperparameters we used are public available\textsuperscript{\ref{toolkit}}.

\subsubsection{Did you report \textit{descriptive statistics} about your results (e.g., error bars around results, summary statistics from sets of experiments), and is it transparent whether you are reporting the max, mean, etc. or just a single run?}
Yes, we indicated that in Section \ref{sec:setup}.

\subsubsection{If you used existing packages (e.g., for preprocessing, for normalization, or for evaluation), did you report the implementation, model, and parameter settings used (e.g., NLTK, Spacy, ROUGE, etc.)?}
Yes, we reported them in Section \ref{sec:superbsg}.

\subsection{Did you use \textit{human annotators} (e.g., crowdworkers) or \textit{research with human subjects}?}
No.

\end{document}